\pdfoutput=1

\documentclass[11pt]{article}

\usepackage[preprint]{acl}

\usepackage{times}  
\usepackage{latexsym}

\usepackage[T1]{fontenc}

\usepackage[utf8]{inputenc}

\usepackage{microtype}

\usepackage{inconsolata}

\usepackage{graphicx}

\usepackage{booktabs} 
\usepackage{times}
\usepackage{latexsym}
\usepackage{amsmath}
\usepackage{amsthm}
\usepackage{amssymb}
\usepackage{graphicx}
\usepackage{float}
\usepackage{siunitx}
\usepackage{enumitem}
\usepackage{subcaption}
\usepackage{tabularx}
\usepackage[most]{tcolorbox}
\usepackage{xcolor}
\usepackage{array}
\usepackage{makecell}

\usepackage{setspace}

\tcbset{
    mybox/.style={
        colback=gray!20, 
        colframe=gray!20, 
        boxrule=0.5mm, 
        arc=2mm, 
        left=2mm, 
        right=2mm, 
        top=1mm, 
        bottom=1mm, 
        width=\linewidth, 
        boxsep=1mm 
    }
}

\newif\ifsubmission
\submissiontrue

\newcommand{\githubLink}{
    \ifsubmission
        http://github.com/[redacted]%
    \else
        http://github.com/daynauth/elo\_paper%
    \fi
}

\theoremstyle{definition}
\newtheorem{definition}{Definition}[section]

\title{Ranking Unraveled: Recipes for LLM Rankings in Head-to-Head AI Combat}

\author{Roland Daynauth, 
  \bf{Christopher Clarke, Krisztian Flautner, 
Lingjia Tang, Jason Mars} \\
University of Michigan \\
\{daynauth, csclarke, manowar, lingjia, profmars\}@umich.edu
  }

\begin{document}
\maketitle

\begin{abstract}






Evaluating large language models (LLMs) is a complex task. Pairwise ranking, where humans compare LLM outputs based on predefined criteria, has become a leading approach. By aggregating these comparisons through algorithms such as Elo, rankings across multiple LLMs can be derived. However, applying ranking algorithms in LLM evaluation presents several challenges. Traditional systems like Elo, designed initially for structured competitions such as chess, often produce inconsistent and unstable rankings due to the dynamic and context-dependent nature of LLM performance. Despite the increasing reliance on these methods, a systematic study of ranking algorithms for LLM evaluation remains lacking. This paper examines the effectiveness of various ranking systems for head-to-head LLM comparisons. We define key principles for robust ranking, conduct extensive evaluations of different ranking algorithms, and analyze their stability, accuracy, and sensitivity to real-world conditions. Our findings offer insights into the limitations of existing approaches and provide guidelines for selecting the most appropriate ranking method based on evaluation objectives and resource constraints.

\end{abstract}

\section{Introduction}
\begin{figure}[t] 
    \captionsetup{font=small}
    \centering
    \includegraphics[width=0.35\textwidth]{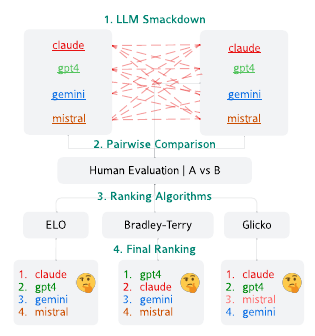}
    \caption{Different ranking algorithms can produce different rankings with the same human evaluation data, making it difficult to determine which algorithm is appropriate for various use cases.}
    \label{fig:elo_diff}
\end{figure}


Rapid progress in large language models (LLMs) has highlighted the critical need for LLM evaluation methodologies, especially for complex generation tasks \cite{guo2023evaluating}. Traditionally, NLP evaluation has relied on benchmarks such as GLUE \cite{wang2019glue}, SuperGLUE \cite{wang2020superglue}, and more recent LLM-focused benchmarks like LM-Eval \cite{eval-harness}. However, a significant limitation of these classical benchmarks is their reliance on established ground truth, making them unsuitable for evaluating more nuanced, diverse, and complex cases where such ground truth is difficult to define or entirely unattainable \cite{chiang2024chatbot}. This problem is particularly evident in complex tasks such as open-ended text generation, conversational dialogue, and creative writing, where human judgment is paramount. Previous work has shown that these benchmark evaluation results do not correlate well with human LLM evaluations \cite{zheng2024judging, irugalbandara2023scaling}.

This shortcoming has led the NLP community to adopt pairwise human evaluation methods \cite{dubois2024alpacafarm, zheng2024judging, chiang2024chatbot}. A widely recognized example is \emph{Chatbot Arena} \cite{chiang2024chatbot}, which directly compares models through "head-to-head combat," where users vote between responses from two competing models. This method aligns with the Elo rating system \cite{elo1978rating}, initially developed to classify chess players according to their historical competition results and relative strengths. Elo and other ranking methods have gained popularity as structured methods incorporating human feedback and preferences to effectively rank models \cite{Dettmers, Wu_Aji_2023, vicuna2023}.

However, as our experiments later demonstrated,  \emph{Elo and other established ranking algorithms can often produce inconsistent and sometimes unreliable results} (Figure~\ref{fig:elo_diff}), generating conflicting ranks. Thus, it is critical to understand how to select the ranking algorithm appropriately. There has been a lack of comprehensive studies across multiple ranking algorithms applied to LLM evaluations. \begin{itemize}
    \item  First, a general framework for such a study has been lacking. It remains unclear which key metrics should be used to evaluate ranking algorithms and the factors that can affect the accuracy of the ranking algorithm.
    \item There has been no comprehensive study comparing multiple ranking algorithms or offering guidelines for selecting the most suitable in the context of LLM evaluations. Although prior work such as \citet{boubdir2023elo} has explored some challenges associated with using Elo for LLM ranking, its focus is limited to Elo. In addition, our experiments show that recommendations from previous work are not always effective in practice.
\end{itemize}

Choosing the best model can significantly impact the user experience of the application that incorporates it. This paper explores the critical issues surrounding ranking LLMs in head-to-head human evaluations. Our contributions are as follows.

\begin{itemize}
    \item We first systemically outline a set of fundamental properties that ranking systems should adhere to when evaluating LLMs.
 
    \item Second, we evaluate the performance of four widely used ranking algorithms (\textit{Elo, Bradley-Terry, Glicko, and Markov Chain}) in two different evaluation scenarios (\textit{Arena Style vs. Controlled Style}), analyzing their ability to maintain transitivity, ensure prediction accuracy, and remain stable across hyperparameter variations. 

    \item Third, we conduct an in-depth analysis of these results, highlighting flaws in the current application of these popular algorithms. Despiteits wide use for LLM ranking, Elo remains unstable even after extensive tuning is applied. 
    

    \item Lastly, we propose systematic guidelines for selecting the most appropriate approach based on the specific characteristics of the evaluation task and the available resources. 

\end{itemize}
To our knowledge, this paper represents the first work that systematically addresses the complexities and intricacies of ranking LLMs across various methodologies. We release all code, data, and models to facilitate reproducibility and further research in this area.

\section{Background}
In this section, we formally define the task of pairwise ranking and briefly describe the suite of algorithms often used for aggregating ranking from pair-wise comparison results. The first three algorithms—Elo \cite{elo1978rating}, Bradley-Terry \cite{bradley1952rank}, and Glicko \cite{glickman1999parameter}—are well-established and widely used in contexts such as sports competitions, while the Markov chain, based on a random walker model, is simpler in design and has demonstrated strong performance in previous studies \cite{Callaghan_Mucha_Porter_2003, Callaghan_Mucha_Porter_2004}.

\subsection{Ranking}
Let $M$ be the set of competition models where $M = \{1, 2, \cdots ,m\}$. For any two models $i$ and $j$ in $M$, the result $S_{ij}$ of a head-to-head match between $i$ and $j$ is defined as $S_{ij} = 1$ when $i$ beats $j$ and $S_{ij} = 0$ when $i$ loses and $S_{ij} = 0.5$ for ties. Let $p_{ij}$ be the probability that the model $i$ beats $j$ in a head-to-head competition. The model $i$ is ranked higher than $j$ if $p_{ij} > 0.5$.

Let $\theta_i \in R$ represent the strength of a model $i \in M$, then it is generally assumed that $p_{ij}$ follows a probability distribution $F$ \cite{david1988method}, such that:

\begin{equation}
    \label{eq:rank}
    p_{ij} = F(\theta_i -  \theta_j)
\end{equation}



\subsection{Elo Rating system}
The Elo rating system was first adopted by the United States Chess Federation (USCF) in 1960 to measure players' strengths \cite{elo1978rating}. For two players $i$ and $j$ with relative strengths $\theta_i$ and $\theta_j$, the probability that $i$ beats $j$, is calculated as follows:

\begin{equation}
   p_{ij} = 1/(1 + 10^{(\theta_j - \theta_i)/400})
\end{equation}

The parameter $\theta_i$ for the model $i$ is also known as the Elo score, updated based on two factors: the outcome of the game $S_{ij}$ and the opponent's Elo score $\theta_{j}$. The score of a player is updated at the end of every match or tournament. In the context of LLM evaluation, we consider a single evaluation run as a tournament and update the score of a model after every head-to-head match. For any match between model $i$ and $j$, the Elo score $\theta_i$ is updated by the function: 
\begin{equation}
\theta^{\prime}_i = \theta_{i} + k\times(S_{ij} - p_{ij})   
\end{equation}

$k$ is a hyperparameter known as the factor $k$ (or the value $k$), which determines the rate at which the Elo score of a player changes based on the outcome of the game. 

\subsection{Bradley-Terry Model}
The Bradley-Terry model \cite{bradley1952rank} was introduced in 1952 as a probability model to predict the result of a paired comparison. The strengths $\theta_i$ for each model can be estimated by solving the likelihood function $\mathcal{L}(\theta)$ such that:

\begin{equation}
    \mathcal{L}(\theta) =  \prod_{i < j}p_{ij}^{y_{ij}}p_{ji}^{y_{ji}}
\end{equation}

where $y_{ij}$ is the number of times that model $i$ beats $j$ and $p_{ij}$ is the probability of $i$ beating $j$. 

\subsection{Glicko Rating system}
The Glicko rating system \cite{glickman1999parameter} was introduced in 1995 as an improvement to Elo. Glicko extends Elo by introducing a secondary parameter $(\sigma)$ known as the \textit{rating deviation}. This parameter measures the reliability of a player's rating score, acting as a confidence interval for the player's rank, decreasing based on the number of games played, regardless of wins or losses. This factor is important given that two players can potentially have the same score in a tournament despite a different number of previous matches. This situation is prevalent in LLM evaluations, as new models are frequently introduced and the number of games played between them vary significantly. In Appendix \ref{sec:glicko_details} we explore the details of the algorithm.









\subsection{Markov Chain}
The Markov Chain Model is a nonparametric ranking algorithm adapted from \citet{Callaghan_Mucha_Porter_2003, Callaghan_Mucha_Porter_2004} and adapted into a Markov Chain Model by \cite{kvam2006logistic} (see Appendix \ref{sec:markov_details} for details). The Markov chain employs a series of random walkers traversing a graph, with each node representing a player/model in the tournament connected by an edge denoting a match between two nodes. Each walker chooses an initial node $i$, staying in the node or moving to an adjacent node $j$ according to a single condition $p$. In a random match between $i$ and $j$, the walker moves to the winning node with a probability of $p$ for $p > 0.5$, a hyperparameter chosen in advance. Every time a walker moves or stays on a node, it represents a vote for that node, and the sum of all votes from every walker represents the rank of that player/model. In practice, the optimal value for $p$ can vary depending on the tournament. In our experiments, we use $p = 0.8$ as the default value similar to \citet{Callaghan_Mucha_Porter_2003}, which achieved stable results with $p$ ranging from $p= 0.75$ to $p = 0.95$.






\section{Desirable Properties of Ranking System} \label{sec:properties}
\label{sec:algo_limit}

The goal of a ranking system is to provide a structured approach to compare and order items based on their performance or quality. In the context of LLMs, an effective ranking system allows researchers, practitioners, and end-users to identify the most capable models for their specific needs. . By simplifying decision-making and optimizing resource allocation, such a system highlights models that are likely to deliver superior performance in the real-world applications. However, implementing a robust ranking system requires addressing several key properties to ensure its validity and reliability. In this section, we identify and discuss three essential properties that a ranking system for LLMs should adhere to: 1) \textit{Transitivity}, 2) \textit{Prediction Accuracy}, and 3) \textit{Sensitivity to Hyperparameters and Battle Conditions}.

\subsection{Transitivity}

Transitivity is a fundamental property for a ranking system to maintain consistency.

\begin{definition}
    \label{def:transitivity}
        For any three models $i, j, k \in M$; their match-ups are \textit{transitive} if $p_{ij} > 0.5, p_{jk} > 0.5$ and $p_{ik} > 0.5$. If $i, j, k$ are transitive, we say that a ranking system \textit{preserves transitivity} if it ranks these models in order $i, j, k$. Transitivity ensures that the ranking is coherent and interpretable.
\end{definition}



Human evaluation data contains transitive and intransitive relationships, whereas ranking algorithms always produce perfectly transitive rankings. Transitive and intransitive relationships are mutually exclusive; however, intransitive relationships are inevitably lost, since ranking enforces transitivity. One of the key strengths of a ranking algorithm lies in its ability to preserve as much of the original transitive structure from human data as possible. Failure to do so can lead to paradoxical rankings, where the ordering of models becomes inconsistent, ultimately diminishing trust in the rating system.

\subsection{Prediction Accuracy}
Like transitivity, prediction accuracy is a critical metric that assesses how well the ranking system forecasts the results in head-to-head matches. It measures the probability that the ranking system correctly predicts the winner between any two LLMs. High prediction accuracy indicates that the ranking system is reliable and can be trusted to make informed decisions about the superiority of the model for a downstream task. For combat-style evaluation in LLM, where assessment is based on subjective human judgment, prediction accuracy acts as an indicator of the ranking system's ability to align with human preferences. This alignment is crucial for tasks like open-ended text generation or conversational agents, where the model's performance is best judged by human preference. Thus, the ranking system should reflect statistical superiority and resonate with qualitative human evaluations to truly enhance the utility and adoption of LLM.

\subsection{Sensitivity to Hyperparameters}

Another critical aspect of evaluating ranking algorithms is understanding their sensitivity to hyperparameters, which can significantly affect the consistency and stability of rankings \cite{boubdir2023elo}. Ranking algorithms such as Elo have been shown to produce drastically different rankings based on the order in which matches are evaluated and the number of permutations \cite{boubdir2023elo}. Since these hyperparameters can influence the behavior and outcomes of the ranking system, it is essential to choose the right settings to ensure reliable and consistent rankings. Sensitivity to hyperparameters means that the ranking system should respond to changes in these settings in a predictable and manageable way. An overly sensitive system might produce drastically different rankings with minor parameter changes, leading to instability and unreliability. In contrast, a system with too little sensitivity might fail to capture essential differences between models. In this work, we consider the impact of two different classes of hyperparameters/conditions: 1) \textit{Algorithmic Hyperparameters}: parameters native to the ranking algorithm itself and 2) \textit{Matchup Distribution}: Arena style vs. controlled style. 

Algorithmic Hyperparameters such as $k$-factor in Elo, $p$-value in Markov Chain, and $\sigma$ in Glicko can significantly affect the rankings produced by these algorithms. Similarly, the distribution of matchups between models can also influence the rankings. In practice, certain models may participate in many more matchups than others, creating an uneven distribution of data. This skews the rankings because models with fewer matches do not have sufficient data to be ranked reliably, while those with more matchups dominate the rankings. The ranking system chosen should be robust to these variations and produce consistent and reliable rankings across different hyperparameters and battle conditions.

\section{Evaluating Ranking Algorithms} \label{sec:evaluation}
Building on the properties discussed in section \ref{sec:properties}, this section outlines a series of experiments designed to assess the quality of traditional ranking systems and their use in LLM rankings. We investigate how each algorithm performs under different conditions and evaluate their effectiveness in producing stable and consistent rankings.


\begin{table}[]
    \captionsetup{font=small}
    \tiny
      \resizebox{\columnwidth}{!}{%
      \begin{tabular}{l|cc}
        \textbf{Algorithm} & \textbf{Arena (\%)} & \textbf{SLAM (\%)}  \\
        \midrule
        Elo           & 68.24        & 52.5               \\
        Markov        & 51.38        & 51.67                \\
        Glicko        & 56.54        & 53.33                 \\
        Bradley-Terry    & \textbf{77.29}        & \textbf{56.67}               \\
      \end{tabular}}
      \caption{Ranking Algorithm Performance on preserving transitivity on Chatbot Arena \& SLAM Dataset.}
      \label{tab:ranking_transitivity}
    \end{table}

\subsection{Datasets}
To evaluate these ranking systems, we use two diverse human evaluation datasets representative of two evaluation styles:

\begin{itemize}

    \item \textbf{Chatbot Arena} \cite{chiang2024chatbot} is a dynamic platform where users evaluate two randomly selected model outputs head-to-head. As newer models are released, they are added to the arena to battle the older models. This data set consists of 57 models and a total of 244,978 match-ups. The distribution of matches varies wildly, with some models having up to 30416 matches or as few as 954. As such, we refer to this dataset as an \textit{Arena-Style} dataset.
    
    \item \textbf{SLAM} \cite{irugalbandara2023scaling} consists of pairwise match-ups between models evaluated using the SLAM tool on domain-specific task questions. Unlike Chatbot Arena, the SLAM platform is tightly controlled with a fixed number of models and a similar number of match-ups between models, thus making it a controlled style dataset. The SLAM dataset consists of 11 models with a total of 2858 match-ups, each model having at most 529 match-ups and at least 501. As such, we refer to this dataset as a \textit{Controlled Style} dataset.
    
\end{itemize}

These two diverse datasets enable us to assess ranking algorithms under varying conditions, offering a comprehensive view of their behavior. To evaluate dataset consistency, we apply Kendall's test \cite{kendall1940method} (see \ref{sec:intrasitivity} for details), which confirms that any inconsistencies in the datasets are statistically insignificant.



\subsection{Preserving Transitivity} \label{sec:competition_trans}

Using the Chatbot Arena and SLAM datasets, we evaluated the ability of each ranking algorithm to preserve transitivity (see \ref{def:transitivity}).  We evaluated the transitive performance of each ranking algorithm by first calculating the number of triples within the data set. Three models $i, j, k$ are part of a triple if $i$ wins the majority of comparisons with $j$, and $j$ wins the majority of comparisons with $k$, and transitivity is considered preserved if this order is maintained in the final ranking. Table \ref{tab:ranking_transitivity} shows the performance of each ranking algorithm under the arena and controlled style evaluation scenarios. We observe that the Bradley-Terry model is the most performant, preserving transitivity in $77.29\%$ and $56.67\%$ of the triples, respectively, compared to its closest competitor, Elo, at $68.24\%$ and $52.5\%$.


Bradley-Terry uses Maximum Likelihood Estimation to calculate each player's strength based on the outcome of all their matches concurrently. This is in contrast to Elo, which updates each player's rating after each match sequentially and is therefore sensitive to the other in which matches are evaluated. This gives Bradley-Terry the advantage in preserving transitivity in larger datasets such as Chatbot Arena. 

\textbf{\small Bradley-Terry consistently outperforms other algorithms in preserving transitivity, achieving the highest rankings across both datasets. }

\begin{table}[]
    \captionsetup{font=small}
    \tiny
    \resizebox{\columnwidth}{!}{%
    \begin{tabular}{l|cc}
            \textbf{Algorithm} & \textbf{Arena (F1)} & \textbf{SLAM (F1)}  \\
            \midrule
            Elo  & \textbf{0.90} &  0.87 \\
            Markov  & 0.77  &  \textbf{0.88}\\
            Glicko  & 0.88 &  \textbf{0.88}\\
            Bradley-Terry & 0.82 & 0.87 \\
        \end{tabular}}
        \caption{Ranking algorithm performance in prediction accuracy (F1 score) on Chatbot Arena \& SLAM Dataset.}
        \label{tab:slam_f1}
    \end{table}

\subsection{Prediction Accuracy} \label{sec:prediction_accuracy}

Next, we evaluate the ability of each ranking algorithm to predict the outcome of unseen LLM matches. We first partition each dataset into a 75/25 train/test split, running each ranking algorithm on the training set to determine the final ranking and strength of each model $\pi$. We then use each ranking algorithm to predict the outcome of the matches in the test set and calculate the F1 score as our performance metric.

\paragraph{Performance on the Arena Dataset}
Due to its uneven distribution of model match-ups, the Arena data set presents an interesting challenge for ranking algorithms to predict future outcomes. When matches between models are unevenly distributed, some algorithms struggle with prediction accuracy. Our results show the Elo model to be the most accurate, achieving an F1 score of $0.90$ compared to the F1 score of $0.88$, $0.82$, and $0.77$ for the Glicko, Bradley-Terry and Markov Chain models, respectively (Table \ref{tab:slam_f1}).

The impact of the match-up distribution on the prediction accuracy is shown in the performance of the Markov chain which is affected by the sparsity of certain model match-ups, leading to lower prediction accuracy, as shown by the poor performance on models such as \textit{deepseek-llm-67b-chat} and \textit{dolphin-2.2.1-mistral-7b} in figure \ref{fig:arena_f1} of the appendix.

Another example of the impact of matchup distribution is shown in the performance of the Bradley-Terry model in the appendix figure \ref{fig:arena_f1}. We observe that the Bradley-Terry system suffers when dealing with "powerful models," where there is a strong imbalance between a model's wins and losses. For example, in the arena dataset, \textit{gpt-4-turbo} had a win/loss ratio of 12288/3979. This skew in model wins/loss results in overestimation of the model's strength as shown by the low F1 score of $0.82$ achieved by \textit{gpt-4-turbo} using Bradley-Terry compared to an F1 score of $0.95$, $0.96$ and $0.92$ from Elo, Glicko, and Markov Chain, respectively. This is known as the `rare events' problem in logistic regression \cite{king2001logistic}. One method of countering this is to use a weighted logistic regression\cite{chiang2024chatbot}. However, we observe that this has a negligible impact on the final result. Furthermore, the bootstrap confidence interval used by Chatbot Arena shows a narrow range of [-2.2, 2.3], suggesting a high confidence in the model ranking. However, this interval does not guarantee how well the model will perform against unseen opponents.

\paragraph{Prediction accuracy on the SLAM dataset}
The distribution of matches between models in the SLAM dataset is tightly controlled so that every reviewer evaluates an equal number of models, resulting in a near uniform distribution of matches. We observe near-identical performance for all models. This result is due to the balanced distribution of match-ups in the dataset, where if a model has the most wins in the competition, its performance is directly reflected in its win rate. We further validate this occurrence by ranking all the models in SLAM by win rate and calculating the correlation between each pair of ranking algorithms as shown in Table \ref{tab:slam_corr}. We observe a high correlation between all methods and the win rate.

\textbf{\small All algorithms perform similarly on the SLAM dataset, but on Arena, Elo leads in accuracy, followed by Glicko. Bradley-Terry, however, is negatively affected by the uneven distribution of the Arena dataset.}



\subsection{Hyperparameter Sensitivity}
\begin{figure*}[h]
    \captionsetup{font=small}
    \centering
    \includegraphics[width=1.0\linewidth]{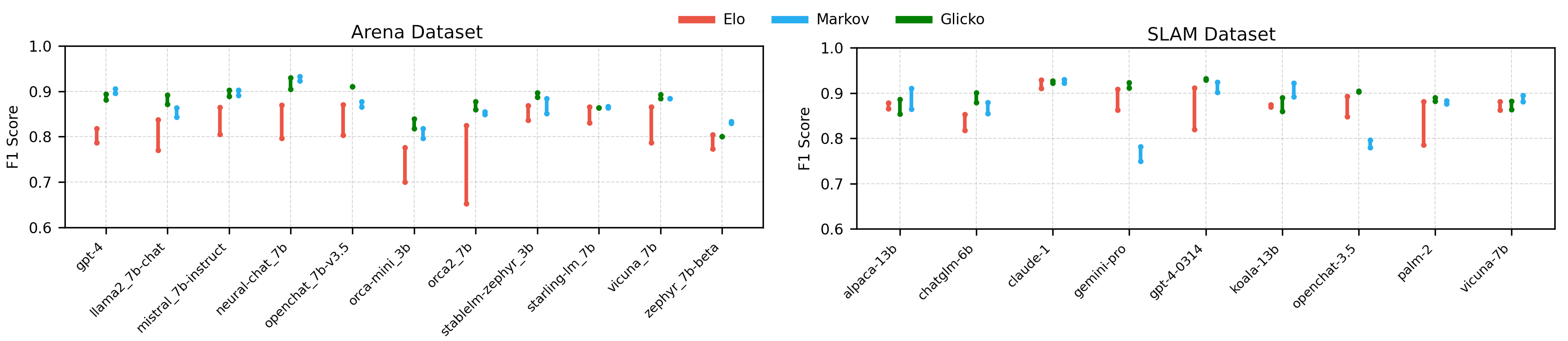}
    \caption{Distribution of F1 scores for the SLAM (left) and ARENA (right) datasets, showing the performance of Elo, Markov, and Glicko algorithms across a subset of models with 100 different hyperparameter settings. The results highlight the volatility of Elo, demonstrating its high sensitivity to hyperparameter changes compared to the more stable performance of Markov and Glicko. }
    \label{fig:f1_scores_boxplot2}
\end{figure*}

To measure the sensitivity of the hyperparameters, we calculate the F1 score of each ranking system as described in Section \ref{sec:prediction_accuracy}, varying each hyperparameter over a range of 100 values. Figure \ref{fig:f1_scores_boxplot2} showcases the range of results produced by Elo, Markov and Glicko across the two evaluation scenarios.


\begin{table}[t]
    \captionsetup{font=small}
    \resizebox{\columnwidth}{!}{%
    \begin{tabular}{l|ccccc}
    \textbf{Method} & \textbf{Elo} & \textbf{Markov} & \textbf{Glicko} & \textbf{Bradley-Terry} & \textbf{Win-Rate} \\
    \midrule
    Elo Rating    & 1.00/1.00 & 0.74/0.89 & 0.86/0.93 & 0.94/0.95 & 0.93/0.91 \\
    Markov  & -         & 1.00/1.00 & 0.76/0.99 & 0.81/0.97 & 0.78/0.98 \\
    Glicko        & -         & -         & 1.00/1.00 & 0.81/0.99 & 0.89/0.99 \\
    Bradley-Terry & -         & -         & -         & 1.00/1.00 & 0.91/0.98 \\
    Win-Rate      & -         & -         & -         & -         & 1.00/1.00 \\
    \end{tabular}}
    \caption{Spearman's Correlation between pairs of ranking algorithms for Arena/SLAM datasets}
    \label{tab:slam_corr}
\end{table}


\paragraph{Elo k-factor}
\label{sec:elo_k_factor}
The $k$ factor in Elo dictates the extent to which the score of a player is influenced by the outcome of an individual match. For example, with large $k$ values, highly ranked players can incur a significant penalty when losing to a player with a lower ranking, while defeating such a player only results in a slight increase in score.  The evaluation of the performance of the model (depicted in Figure \ref{fig:f1_scores_boxplot2}) at varying levels of hyperparameter \( k \) shows marked sensitivity, which emphasizes the need for meticulous tuning. Elo shows the largest variation across both datasets, and is less stable, particularly for the smaller SLAM dataset. Through our tuning of this parameter, we determined that the ideal $k$ factor for Elo is less than the standard 32 commonly used in previous research \cite{Dettmers, Wu_Aji_2023, vicuna2023}. This reduced convergence rate can be attributed to the complexities of human preferences as opposed to the typical assessment of real-world games, which rely on the differing abilities of opposing teams.


\paragraph{Markov Chain: $p$ Score}
The $p$ score in the Markov Chain determines the extent to which the overall ranking is influenced by each model's win rate, with values closer to $1$ indicating greater influence compared to those closer to $0.5$. We observe that Markov shows stable performance on the controlled SLAM datasets, performing as well as Glicko (Figure \ref{fig:f1_scores_boxplot2}), however, the algorithm struggles on the large ARENA dataset due to its over reliance on win-rates for calculating ranks. When optimizing the value of this parameter, we find that the optimal $p$ value for the Markov chain on the SLAM dataset is larger than the Arena dataset, indicating that the model's strength is greatly influenced by its overall win rate.


\paragraph{Glicko: Rating deviation}
For Glicko, the ranking deviation $(\sigma)$ is a measure of consistency or variability in a player's performance over time and determines how flexible a player's rating can be. If $\sigma$ is initially set too high, the system will interpret the early game results as more variable or uncertain. This leads to larger rating adjustments, making the player's rating fluctuate more. In contrast, setting $\sigma$ too low initially restricts the system from adjusting the rating freely. As a result, players, especially newcomers, may take much longer to reach a rating that accurately reflects their true skill level. This is particularly impactful in fast-changing or skill-evolving games, where a low initial variability setting could lead to underestimation of skill. 
We observe that Glicko's prediction performance is consistent across both datasets, showing that its performance is less impacted by changes in its hyperparameters compared to Elo and Markov Chain. We theorize that this is due to the large number of games played for each model in both scenarios, allowing the system to adjust the rating deviation dynamically.

\textbf{\small  Elo shows high sensitivity to hyperparameter whereas Glicko demonstrates greater stability and accuracy which suggests that Glicko is a more reliable alternative for LLM evaluation.}

\section{Recommended Best Practices}

\begin{table*}[htbp]
\captionsetup{font=small}
\centering
\tiny
\begin{tabularx}{\textwidth}{|l|X|X|X|X|}
\hline
\textbf{Criteria} & \textbf{Elo} & \textbf{Markov} & \textbf{Glicko} & \textbf{Bradley-Terry} \\ \hline
\textbf{Transitivity} & 
Moderate performance on smaller datasets, better than average performance on larger datasets. & 
Preserves transitivity but sensitive to sparse and uneven matchups. & 
Good transitivity preservation in both types of datasets. & 
Generally good at maintaining transitivity, in smaller controlled and larger uncontrolled datasets. \\ \hline

\textbf{Hyperparameter Sensitivity} & 
Highly sensitive to K-factor and permutation count; performance fluctuates with changes. & 
Sensitive to the p-value, with rankings varying significantly based on its setting. & 
 Incorporates rating deviation, making it less sensitive than Elo, but hyperparameter tuning is still necessary. & 
Does not use any hyperparameters. \\ \hline

\textbf{Prediction Accuracy} & 
Moderate prediction accuracy but inconsistent with small sample sizes or highly uneven matches. & 
Moderate prediction accuracy but highly dependent on matchup distribution. & 
High prediction accuracy due to consideration of the uncertainty in the ratings (rating deviation). & 
Higher prediction accuracy in controlled datasets but suffers in uneven matchups. \\ \hline

\textbf{Recommended Use Cases} & 
Not recommended for small, unevenly distributed datasets; may perform better in larger datasets with consistent matchups. & 
Applicable where dataset size is small and even, but sensitive to distribution; effective if properly tuned in moderately structured environments. & 
Best suited for large, unevenly distributed datasets. Able to manage uncertainty and prevent low matchup models from being ranked too high. & 
Ideal for small, unevenly distributed datasets. Also effective for large, evenly distributed datasets where interpretability and scalability are valued. \\ \hline
\end{tabularx}
\caption{Consideration for Ranking Algorithm Selection based on Desired Ranking Properties}
\label{tab:ranking_algorithms_comparison}
\end{table*}

In sections \ref{sec:properties} and \ref{sec:evaluation}, we outlined the desired properties of ranking systems and presented experiments assessing the performance of these systems in achieving these criteria. Based on these observations, this section puts these results into perspective and provides practical recommendations for selecting the most appropriate ranking algorithm. We outline which algorithms perform best under specific circumstances, offering guidance on their optimal use depending on the structure of the dataset, the matchup distribution, and the model evaluation goals. These recommendations, summarized in Table \ref{tab:ranking_algorithms_comparison}, aim to improve the reliability and applicability of classification systems in real-world LLM evaluations.

\subsection{Challenges with Elo}
An important result in this paper is that Elo, even with $>$ 1,000 permutations, struggles to achieve stable rankings \ref{sec:elo_instability}. This indicates that Elo's dependency on permutations for ranking stabilization is inadequate, challenging previous studies \cite{boubdir2023elo}. Combined with the high sensitivity to hyperparameters and the average performance in preserving transitivity. \textbf{We do not recommend ELO as an algorithm to rank LLM performance.}

\begin{tcolorbox}[mybox]
\small 
\textbf{Best Practice 1.}
Elo's high sensitivity to changes in its \(K\) factor and instability (even with large permutations) make it unsuitable for efficient ranking of LLM performance in pairwise evaluation (Sections \ref{sec:elo_k_factor} and \ref{sec:elo_instability}). 
\end{tcolorbox}

\subsection{Choose Bradley-Terry for Small Controlled Datasets}
In section \ref{sec:prediction_accuracy}, we demonstrate that ranking algorithms exhibit varied performance based on battle conditions. When matches are evenly distributed, the ranking algorithms produce similar results. However, a heavily skewed distribution can negatively impact some methods such as Elo and Bradley-Terry. In this scenario, where matchups are balanced and the number of matches is small, objective findings show that Bradley-Terry is best due to its ability to preserve transitivity while also maintaining comparable prediction accuracy.

\begin{tcolorbox}[mybox]
\small 
\textbf{Best Practice 2.}
Bradley-Terry's simplicity and lack of hyperparameters make it ideal for controlled evaluation settings, although its reliance on sufficient matchup data limits its effectiveness in unevenly distributed scenarios.
\end{tcolorbox}

\subsection{Choose Glicko for large uneven datasets}
For larger datasets, especially those with uneven matchup distributions, the Bradley-Terry model encounters difficulties due to the rare event problem. This problem occurs when new models with high win rates, participate in fewer matchups than well-established models. As a result, these new models can be ranked disproportionately higher based on limited data, leading to biased rankings. In this scenario, the Glicko system provides a superior solution. Glicko utilizes a rating deviation parameter to account for uncertainty in a model's ranking due to limited matchups. This allows the system to adjust the rankings more conservatively for models with fewer comparisons, preventing them from being ranked too high based only on a few matches. As data sets become larger and more uneven, Glicko's ability to incorporate rating deviation ensures that the rankings accurately reflect the true performance throughout the dataset, even when new models appear with limited match histories. This flexibility makes Glicko particularly effective for large-scale LLM assessments where uneven matchup distributions are prevalent.

\begin{tcolorbox}[mybox]
\small 
\textbf{Best Practice 3.}
Glicko is effective for large data sets with uneven matches due to its rating deviation parameter, which manages uncertainty and prevents models with few matches from being overrated.
\end{tcolorbox}

\subsection{Bradley-Terry for other Dataset Types}
For small, uneven datasets, Bradley-Terry's simplicity and ability to model pairwise comparisons make it a robust choice, even when the data distribution is skewed. It does not require parameter estimation, which allows it to handle small datasets efficiently without the need for complex adjustments or hyperparameter tuning, making it ideal for situations where uneven distributions might otherwise require additional preprocessing.

In the case of large, evenly distributed datasets, Bradley-Terry is effective because of its scalability and ease of interpretation. For large data sets where the distribution of results is balanced, Bradley-Terry's ability to produce clear, interpretable rankings without adding excessive computational overhead is a significant advantage. Although more complex models such as Glicko can offer finer granularity in large, uneven data cases, Bradley-Terry’s efficiency and interpretability make it a viable alternative when computational simplicity and transparency are prioritized.

\begin{tcolorbox}[mybox]
\small 
\textbf{Best Practice 4.}
Bradley-Terry's scalability and interpretability make it a viable choice for large, evenly distributed datasets, offering clear rankings with minimal computational overhead.
\end{tcolorbox}

\section{Related Work}
Pairwise ranking has recently emerged as a popular method for evaluating LLMs, as it allows for direct comparisons between models and provides a more nuanced assessment of their performance than traditional metrics \cite{chiang2024chatbot}. This evaluation involves presenting human judges with pairs of model outputs and asking them to select the better one. The outcomes of these pairwise comparisons are then aggregated into final rankings using ranking algorithms such as Elo \cite{elo1978rating}. Despite their widespread use, there has been limited research on the applicability of these ranking algorithms to LLM evaluation. Works such as \cite{boubdir2023elo} have critically analyzed the compatibility of Elo with LLM evaluations by examining how well it adheres to reliability and transitivity. However, Elo is not the only ranking algorithm used in LLM evaluation. There exists a number of other algorithms that have been proposed to rank in different contexts, such as the Bradley-Terry model \cite{bradley1952rank} and Glicko \cite{glickman1999rating}, which have been shown to outperform Elo in certain scenarios, but their performance in the context of LLM evaluation has not been systematically evaluated. However, several factors affect the performance of these algorithms, such as the distribution of the data, the number of models that are compared, and the number of pairwise comparisons. As such, there is a need for a systematic comparison of ranking algorithms in the context of LLM evaluation to identify the most suitable algorithm for different scenarios which this paper aims to address.

\section{Conclusion}
In this paper, we explore the application of several ranking algorithms in the context of evaluating LLMs. Through our experiments and analysis, we identify significant limitations in the ability of these algorithms to maintain stable and accurate ranking across varying dataset types and distribution patterns. To combat these challenges, we introduce a set of guidelines for selecting the most appropriate ranking system based on the specific characteristics of the evaluation task and available resources. We believe that our findings and recommendations will be valuable to researchers and practitioners working in the field of LLM evaluation, and we hope that our work will inspire further research in this area.

\section{Limitations}
\label{sec:limitations}

While our study comprehensively explores the ranking of Large Language Models (LLMs) through head-to-head combat, several limitations warrant discussion.

\paragraph{Scalability Constraints}
One significant limitation arises from the scalability constraints inherent in pairwise evaluations. As the number of LLMs in the ecosystem expands, the number of required comparisons grows quadratically. This introduces computational challenges and resource constraints that may limit the feasibility of exhaustive head-to-head evaluations, particularly for larger sets of models.

\paragraph{Human Feedback Variability}
Another concern is the variability in human feedback, on which our rankings are heavily based. Human judgments can be subjective and influenced by numerous factors, including the individual's background, expertise, and contextual understanding. This variability can introduce noise into the ranking system, affecting the evaluations' stability and reliability.

\bibliography{main}

\newpage
\appendix

\onecolumn

\section{Appendix} \label{sec:appendix}
\subsection{Glicko Algorithm Details}
\label{sec:glicko_details}
For model $i \in M$, the rating and the rating deviation are updated as:

\begin{equation}
\theta_i^{\prime} = \theta_i + \frac{q}{\frac{1}{\sigma^2_i} + \frac{1}{d^2}}\sum_{j=1}^{J}{g(\sigma_j)(S_{ij} - p_{ij})}
\end{equation}

\begin{equation}
\quad \sigma_{i}^{\prime} = \left(\frac{1}{\sigma^2_{i}} + \frac{1}{d^2}\right)^{-1}
\end{equation}

where $q = \log(10)/400$, $J$ is the set of opponents for $i$ and $g(\sigma) = 1 / \sqrt{1 + 3q^2\sigma^2/\pi^2}$. 

Given the rating $\theta_i$ and $\theta_j$ of model $i$ and $j$, the expected outcome $p_{ij}$ is calculated as\footnote{For $\sigma = 0$, $p_{ij}$ defaults to the classic Elo rating system.}:

\begin{equation}
    p_{ij} = \frac{1}{1 + 10^{-g(\sigma_j)(\theta_i - \theta_j)/400}}
\end{equation}

\begin{equation}
     \quad d^2 = \left(q^2\sum_{j = 1}^{J}g(\sigma_{j})^2p_{ij}(1 - p_{ij})\right)^{-1}
\end{equation}

where $d^2$ is the scaling factor for the system's certainty.

\subsection{Markov Chain Details}
\label{sec:markov_details}
The Markov Chain Model was originally developed as a random walker model \citet{Callaghan_Mucha_Porter_2003, Callaghan_Mucha_Porter_2004} where ranks were calculated by solving a series of differential equals. This method was adapted into a Markov Chain Model by \cite{kvam2006logistic}. When expressed as a Markov chain, the elements $T_{ij}$ of the transition matrix $T$ are the probabilities that a walker moves from model $i$ to $j$ in a single time step, with the initial values computed as follows:

 \begin{equation}
     t_{ij} = \frac{1}{N_i}\left[w_{ij}(1-p) + l_{ij}p\right], \quad \forall j\ne i
 \end{equation}

\begin{equation}
    t_{ij} = \frac{1}{N_i}\left[W_ip + L_i(1 - p)\right]
\end{equation}

where $W_i, L_i$ are the number of matches won by model $i$ and $w_{ij}, l_{ij}$ are the number of games won by model $i$ against model $j$.

The steady-state probability vector $\pi = [\theta_i]$ represents the final ranking of each team $i \in M$ such that $\pi T = \pi$ and $\sum_{i = 1}^{m} \theta_i = 1$.

\subsection{F1 scores per model}

In this paper, F1 scores are used as an evaluation metric to assess the consistency and predictive performance of the ranking algorithms. The F1 score is calculated based on the precision and recall of the correctly predicted pairwise outcomes compared to the observed outcomes in the dataset. In the following, we detail the methodology for computing the F1 scores:

Let the data set consist of $m$ pairwise matches between models. For each pair $(i, j)$, $p_{ij}$ is the probability that the model $i$ beats $j$ in a random match. If model $i$ plays $M_i$ matches then $E_i$, the expected number of games won by model $i$ is $E_i = \left \lfloor{M_ip_{ij}}\right \rfloor $ and the actual number of games won is given by $A_i$.

The Precision and Recall are calculated as:

\[
    \text{precision} = \frac{|E_i \cap A_i|}{|E_i|}, \quad \text{recall} = \frac{|E_i \cap A_i|}{|A_i|}
\]

The F1 score is calculated as the harmonic mean of precision and recall:
\[
\text{F1 Score} = 2 \cdot \frac{\text{Precision} \cdot \text{Recall}}{\text{Precision} + \text{Recall}}.
\]

The total F1 score for a ranking algorithm is computed by averaging the F1 scores for all pairwise matches \( (i, j) \):
\[
\text{F1 Score}_{\text{Overall}} = \frac{1}{|M|} \sum_{(i, j) \in M} \text{F1 Score}_{(i, j)},
\]
where \( M \) is the set of all pairwise matches.

\begin{figure*}[h] 
    \centering
    \includegraphics[width=0.9\textwidth]{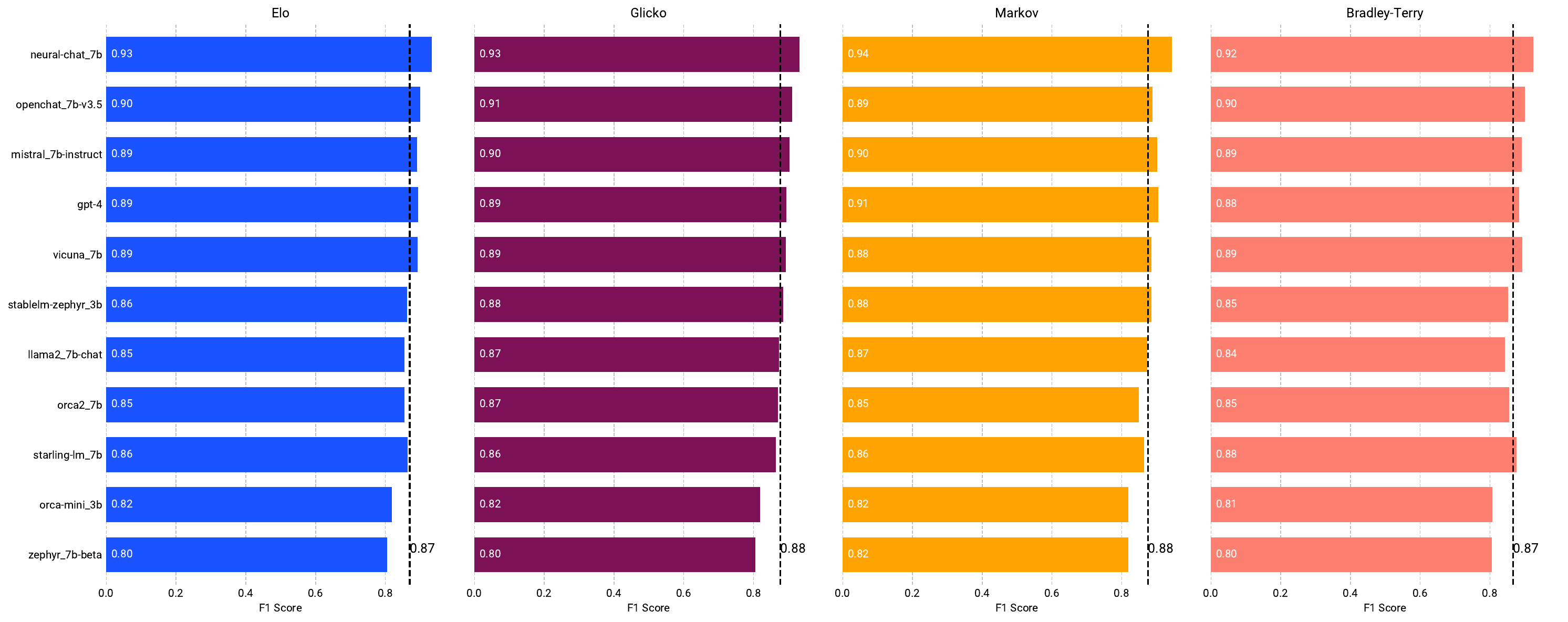}
    \caption{F1 scores for all models in the SLAM dataset. }
    \label{fig:slam_f1}
\end{figure*}

\begin{figure*}[h] 
    \centering
    \includegraphics[width=0.9\textwidth]{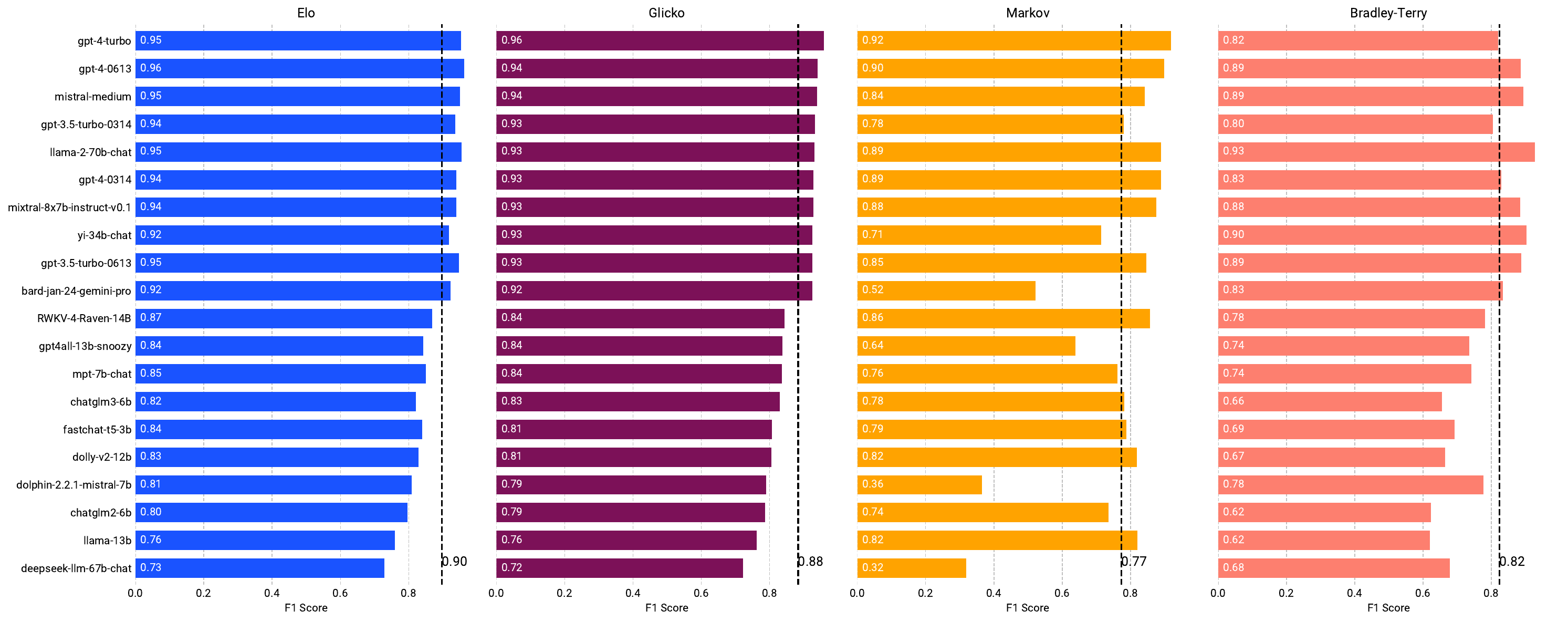}
    \caption{F1 scores of the top ten (10) highest and lowest ranked models using the Elo rating system on the Chatbot Arena dataset. In general, Elo provides the best prediction accuracy by achieving an F1 score of $0.90$. }
    \label{fig:arena_f1}
\end{figure*}

Figures \ref{fig:slam_f1} and \ref{fig:arena_f1} show the F1 scores for both the SLAM and Arena datasets, and also their overall F1 score.

\label{sec:prediction}

\subsubsection{Experimental Setup}
To evaluate the predictive accuracy of the ranking algorithms, each data set was split into a 75\% training set and a 25\% test set. 

\subsubsection{Procedure} The ranking algorithms were applied to the training set to compute model rankings based on pairwise matchups. Using these rankings, the algorithms predicted the outcomes of matchups in the test set, expressed as the number of games each model was expected to win. These predictions were compared with the actual results of the match and the precision, recall and F1 scores were calculated for each match. The F1 scores, aggregated across all matches in the test set, quantified the alignment between the predicted and observed results, providing a measure of the ability of each algorithm to generalize to unseen data.

\subsection{Elo's rank at $k = 3$ vs $k=5$}\label{sec:appendix_a2}
To analyze the sensitivity of Elo rankings to hyperparameter settings, we conducted experiments using two versions of Elo, with \( K = 3 \) and \( K = 5 \), on the SLAM dataset. 

\subsubsection{Experimental Setup}
For each \( K \)-factor, we evaluated Elo rankings under varying numbers of permutations: \( 1, 10, 100, \) and \( 1000 \). Permutations involve randomizing the order of matchups, a strategy commonly employed to reduce the effect of matchup order on final rankings and improve stability.

\subsubsection{Procedure}
\begin{figure*}[ht]
\captionsetup{font=small}
    \centering
    \begin{subfigure}[b]{0.49\textwidth} 
        \captionsetup{font=small}
        \centering
        \includegraphics[width=\textwidth]{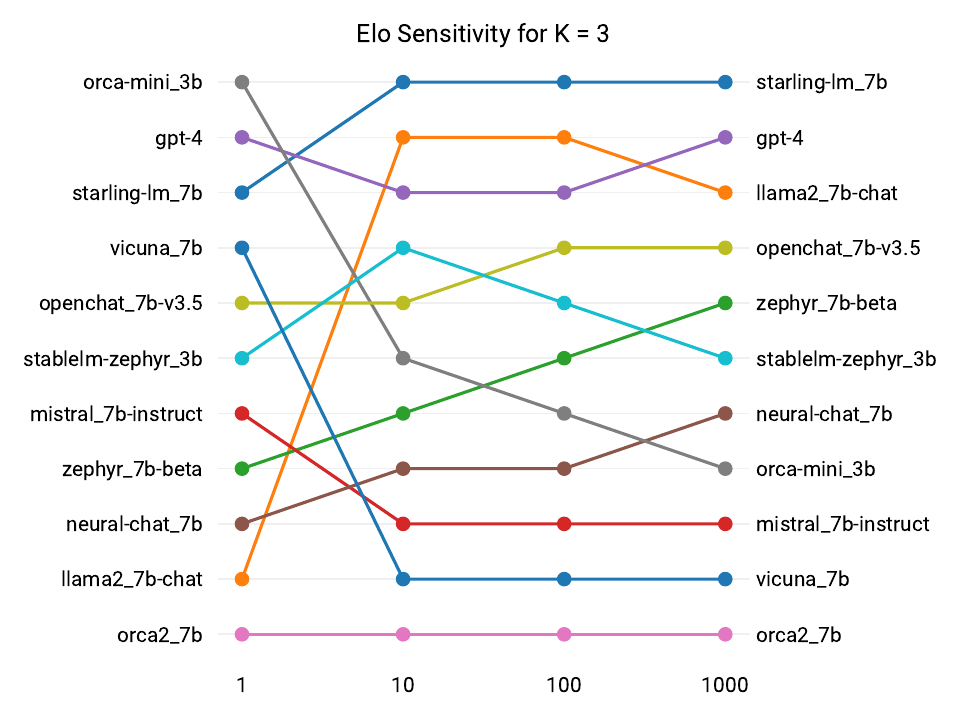}
        \caption{Ranks produced by Elo for $k =3$}
        \label{fig:elo_s_3}
    \end{subfigure}
    \hfill
    \begin{subfigure}[b]{0.49\textwidth}
    \captionsetup{font=small}
        \includegraphics[width=\textwidth]{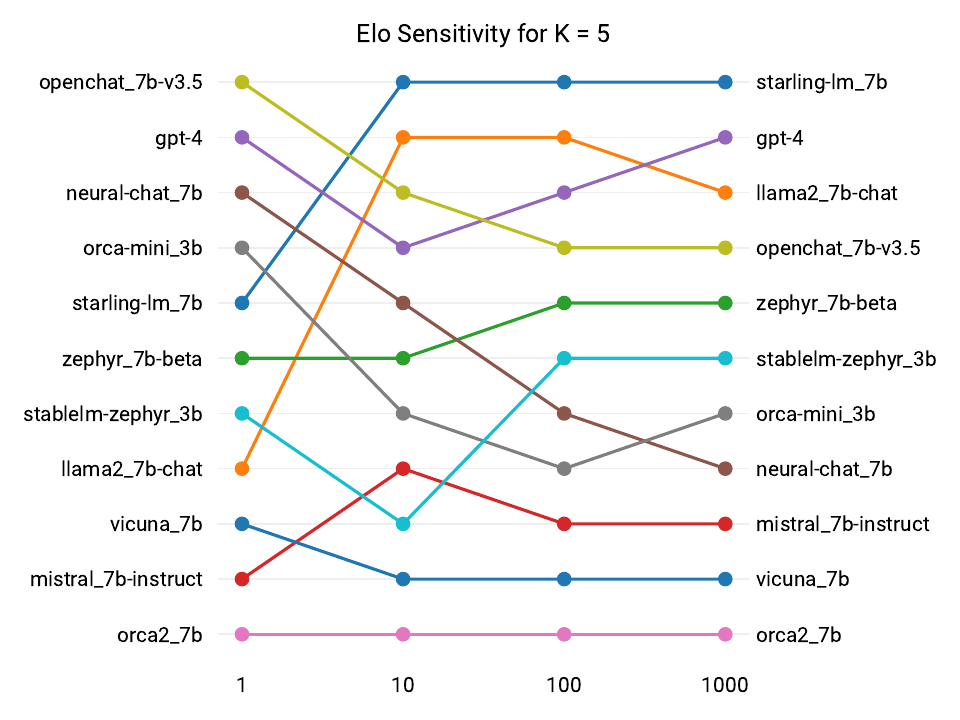}
        \caption{Ranks produced by Elo for $k =5$}
         \label{fig:elo_s_5}
    \end{subfigure}

    \caption{Elo produces different ranks based on the value of the hyperparameter $k$. Increasing the number of permutation can lead of more stable ratings, however, model ranks may still be unstable as is the case with \textit{orca-min\_3b} and \textit{neural-chat\_7b}.}
    \label{fig:elo_k_value_comparison}
\end{figure*}

For each \( K \) factor (\( K = 3 \) and \( K = 5 \)), Elo rankings were calculated using the raw SLAM matchup data without permutations. Matchup orders were permuted \( 1, 10, 100, \) and \( 1000 \) times for each factor \( K \), with Elo rankings recalculated after each set of permutations. Our findings (shown in figure \ref{fig:elo_k_value_comparison}), show volatility of Elo's ranking for each of the $k$ value.

\subsection{Elo's Instability at higher permutation}
\label{sec:elo_instability}
To analyze the performance of Elo at higher permutations, we conducted two separate experiments on the SLAM dataset.

\subsubsection{Experimental Setup}
For a single \( K \) factor of 25, we evaluated Elo rankings by performing two separate evaluations of each \(10,000\) permutation using the strategy of the previous section. 

\subsubsection{Procedure}
\begin{figure*}[h] 
    \centering
    \includegraphics[width=0.9\textwidth]{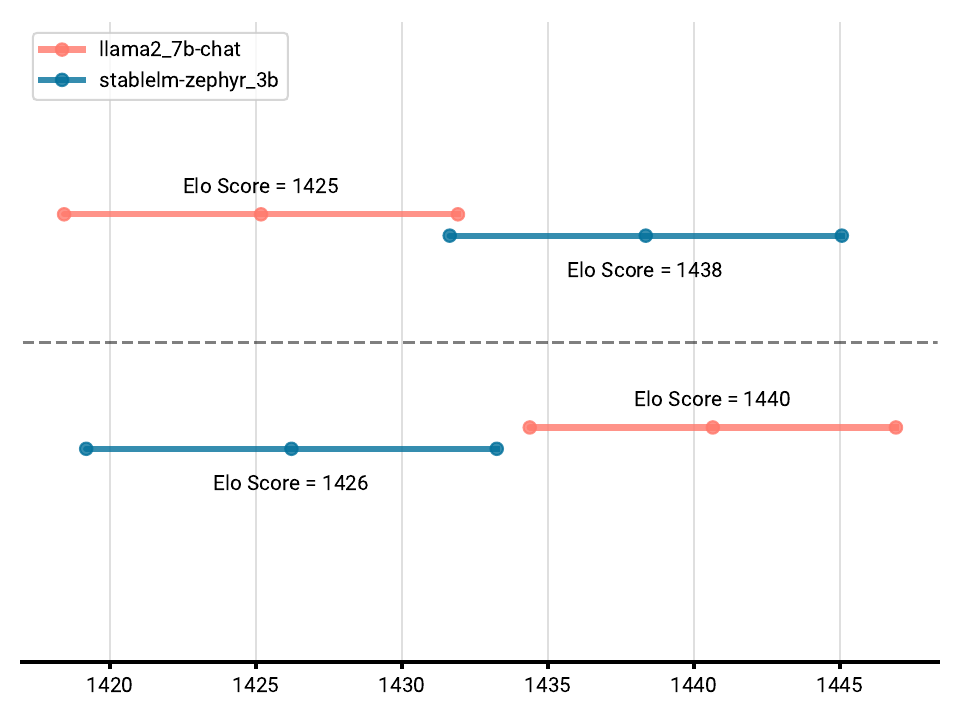}
    \caption{Two separate ranks produced by ELO for permutations of 10,000 each. }
    \label{fig:elo_perm_diff}
\end{figure*}

We observed the ranks produced by each experiment, showing that even with a large permutation, ELO can produce two different ranks for the same two models\ref{fig:elo_perm_diff}.

\label{sec:prediction}

\subsection{Evaluating the consistency of crowd-workers evaluators}
\label{sec:intrasitivity}
In this section, we recommend a method to evaluate the consistency of crowd-worker evaluations based on the Analytic hierarchy process (AHP) practices. We begin by formally defining cycles in a pairwise ranking dataset.
\begin{definition}
    \label{def:rps_cycle}
    For models $i_1, i_2, \cdots i_{k} \in M$, if $p_{i_1i_2} > 0.5, p_{i_2i_3} > 0.5, \cdots, p_{i_{k}i_1} > 0.5$, then their games is said to be \textit{cyclic}. A cycle of length three is called a Rock, Paper, Scissors (RPS) cycle.
\end{definition}

For any three models $i, j,$ and $k$, it is expected that if a reviewer prefers $i$ over $j$ and $j$ over $k$, they prefer $i$ over $k$ (transitivity). However, this may not always be due to personal preference, evaluation error, or evaluator negligence. If the transitive property does not hold, we say that the three models are intransitive and the evaluation is inconsistent. 



Measuring the consistency of pairwise ranking falls under the domain of AHP. According to AHP theory, if the number of RPS cycles remains small, it reflects the complexity of the ranking problem rather than the judges' complacency. However, a high number of cycles represents a low quality of judgments.

According to \cite{kendall1940method}, the expected number of RPS cycles for a pairwise comparison dataset of $m$ items follow a $\chi^2$ distribution, and if the actual number of RPS cycles is significantly less than the expected number, we can be reasonably confident that the dataset is consistent.

The number of RPS cycles can be calculated by first generating the win rate matrix $P$ as defined in section \ref{sec:competition_trans}. This is followed by the adjacency matrix $A$, where $a_{ij} = 1$ if $p_{ij} > 0.5$ and 0 otherwise. Finally, the total number of RPS cycles is calculated as  $\rho = 1/3\sum_{i = 1}^M a^{*}_{i,i}$, where $a^{*}_{ii}$ are diagonals of $A^3$\cite{10.3390/a15050152}.

The $\chi^2$ statistic is calculated as:
\begin{equation}
    \chi^2 = \frac{8}{k-4}\left\{ \frac{1}{4} \binom{k}{3} - d - \frac{1}{3} + \rho\right\}
\end{equation}
\begin{equation}
    v= \frac{k(k - 1)(k - 2)}{(k-4)^2}
\end{equation}

where $v$ is the degrees of freedom.

The RPS count and $\chi^2$ statistic for both the SLAM and Arena datasets are shown in table \ref{tab:chi}. At a p-value close to 0, we can be confident that inconsistencies are not due to the quality of the judges.


\begin{table}[htbp]
\tiny
    \centering
    \caption{Non-Transitivity statistic for the two datasets that were used. The p-value is calculated based on the $\chi^2$ estimated via Kendal's Chi-squared approximation to the circular triad distribution. }
    \small
    \begin{tabular}{|c|c|c|c|}
        \hline
        Dataset & RPS Count & $\chi^2$ & p-value \\
        \hline
        SLAM & 19 & 45.06 & 0.001 \\
        Arena & 393 & 1107.25 & 0.000 \\
        \hline
    \end{tabular}
    \label{tab:chi}
\end{table}

\end{document}